# Apprenticeship Learning for Model Parameters of Partially Observable Environments

**Takaki Makino** MAK@SAT.T.U-TOKYO.AC.JP
Institute of Industrial Science, the University of Tokyo, 4-6-1 Komaba, Meguro-ku, Tokyo 153-8505 Japan

**Johane Takeuchi** JOHANE.TAKEUCHI@JP.HONDA-RI.COM
Honda Research Institute Japan Co., Ltd., 8-1 Honcho, Wako-shi, Saitama 351-0188, Japan

## Abstract

We consider apprenticeship learning — i.e., having an agent learn a task by observing an expert demonstrating the task — in a partially observable environment when the model of the environment is uncertain. This setting is useful in applications where the explicit modeling of the environment is difficult, such as a dialogue system. We show that we can extract information about the environment model by inferring action selection process behind the demonstration, under the assumption that the expert is choosing optimal actions based on knowledge of the true model of the target environment. Proposed algorithms can achieve more accurate estimates of POMDP parameters and better policies from a short demonstration, compared to methods that learns only from the reaction from the environment.

## 1. Introduction

Learning from Demonstration (LfD) is a framework for learning to perform a complex task by observing demonstration (task execution) by an expert (Argall et al., 2009). LfD is particularly useful for domains where the expert knowledge of the domain is limited or difficult to represent, because demonstrations are much easier than designing a controller for the task.

Apprenticeship Learning via Inverse Reinforcement Learning (Abbeel & Ng, 2004), which is an application of LfD for reinforcement learning, is an algorithm that learns the reward function of the environment under the assumption that the expert is trying to maximize the reward. The idea is that, although reinforcement learning can produce an optimal policy with respect to a given reward function, designing a reward function that captures the desired task behavior is not always obvious and requires expert knowledge of the domain. Moreover, learning the reward function from demonstration requires much less amount of demonstration compared to learning the policy directly from the demonstration, because the reinforcement learning combines the reward function with the environment model for optimizing the policy for future rewards. Inverse reinforcement learning is successfully applied to tasks where the environment is fully observable, including aerobatic helicopter flight (Abbeel et al., 2010), robot hand control (Boularias et al., 2011) and prediction of linguistic structures (Neu & Szepesvári, 2009). Inverse reinforcement learning in partially observable environments when an exact model is available has also been studied (Ziebart et al., 2010; Henry et al., 2010; Choi & Kim, 2011).

However, the design bottleneck is not limited to the reward function. In many tasks, how to model the environment is not obvious as well, and requires expert knowledge of the domain, especially when the environment is partially observable. For example, dialogue system tasks are often represented as a Partially Observable Markov Decision Process (POMDP) in which the user's mental state is situated as a hidden state (Williams et al., 2005; Kim et al., 2008; Meguro et al., 2010), but designing such a model requires a considerable amount of work by domain experts, such as annotating dialogue corpus. Thus, there is a need for a way to estimate uncertain parameters of an environment model from non-annotated demonstration data.

One obvious way to estimate environmental parameters from the demonstration is to extract the environmental reaction to the expert's action (Thomson et al.,

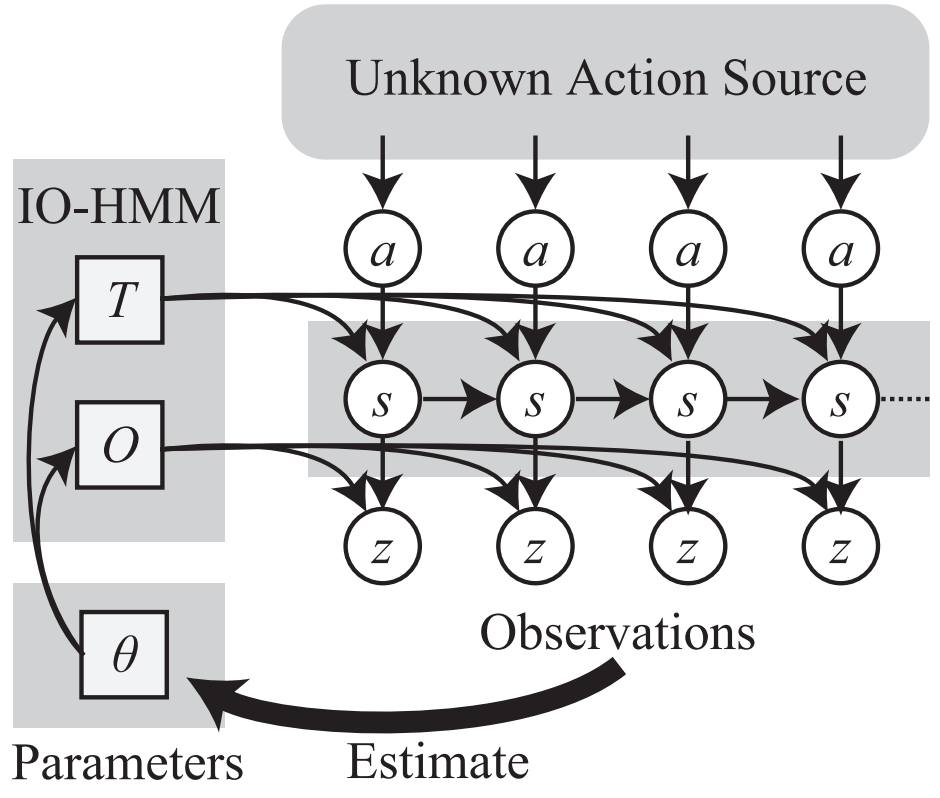

(a) Parameter estimation from environmental response within demonstration.

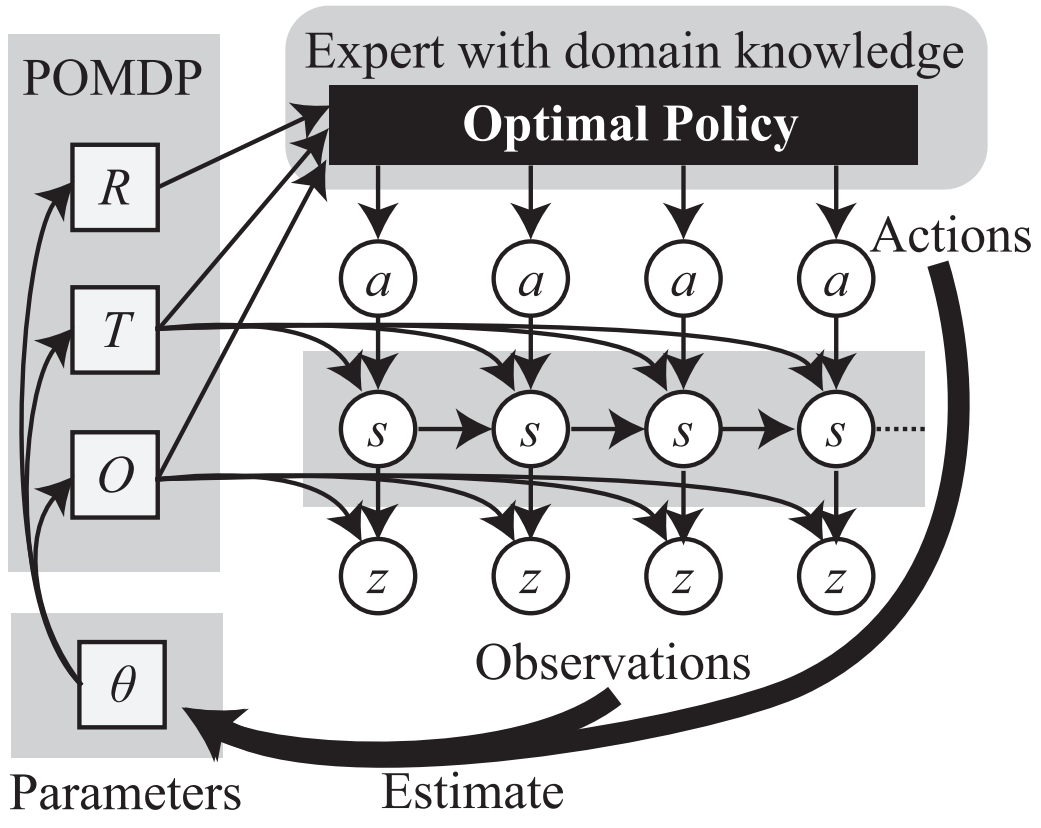

(b) Parameter estimation by apprenticeship learning (proposed).

*Figure 1.* Comparison of existing and proposed methods

2010). In case of POMDP environment, This reduces the problem into a parameter estimation of an Input-Output Hidden Markov Model (IO-HMM) (Bengio & Frasconi, 1996) (Fig. 1(a)). However, this approach assumes nothing about the demonstrator, and it is applicable to cases where the demonstration is generated from the learning agent or even from a naive random policy. Our claim is that demonstration by an expert contains much richer information about the environment that comes from the expert's knowledge, and by extracting this information, we can reduce the burden of designing a model suitable for the task.

Our proposal is to apply the framework of apprenticeship learning to estimate uncertain parameters of the environment (Fig. 1(b))[1] . Assuming that the expert's behavior is based on a stochastic optimal policy with knowledge of the perfect POMDP model for the target environment, we can extract the expert's knowledge regarding the POMDP parameters from his demonstration. The extracted information of expert knowledge can be combined with IO-HMM estimation from the environmental response, to provide a better estimate of the POMDP parameters.

We present two straightforward estimation algorithms, maximum a posteriori (MAP) estimator and posterior sampler by Markov chain Monte Carlo (MCMC), combined with planning algorithms to achieve model-parameter apprenticeship learning In the experiments with short demonstrations, we show that our algorithms can achieve more accurate estimates of POMDP parameters and better policies than can existing methods based on IO-HMM estimation.

[1]We can use this approach for any environment model, such as fully-observable MDPs. However, the most effective cases are for POMDPs because they are hard to be learned from environmental reactions, so we focus on POMDPs in this paper.

## 2. POMDP and its Parameterization

An Input-Output Hidden Markov Model (IO-HMM) (Bengio & Frasconi, 1996) is a framework for representing environments consisting of hidden states, inputs (actions that may affect the states), and outputs (observations from the states). Formally, an IO-HMM is defined as a tuple $\langle S, A, Z, T, O, \boldsymbol{b}_0 \rangle$, where $S$ is the finite set of states, $A$ is the finite set of actions, $Z$ is the finite set of observations, $T$ is the state transition function such that $T(s, a, s')$ denotes probability $P(s'|s, a)$ of changing to state $s'$ by taking action $a$ at state $s$, $O$ is the observation function such that $O(a, s, z)$ denotes probability $P(z|a, s)$ of perceiving observation $z$ as a result of taking action $a$ and arriving in state $s$, and $\boldsymbol{b}_0$ is the vector of initial state distribution such that $b_0(s)$ denotes the probability of starting in state $s$.

Since the true state is hidden, we construct a belief about the state. We denote belief by a vector $\boldsymbol{b}$ where $b(s)$ denotes the probability that the state is $s$ at the current time step. The following update formula can be used to calculate the belief $\boldsymbol{b}_z^a$ for the next time step from the belief at the current time step, given the action $a$ at the current time step and the observation $z$ at the next time step:

$$b_z^a(s') \propto O(a, s', z) \sum_s T(s, a, s') b(s) \quad . \tag{1}$$

A partially observable Markov decision process (POMDP) is a formulation of an action selection problem on an IO-HMM. A POMDP is defined as a tuple $\mathcal{P} = \langle S, A, Z, T, O, \boldsymbol{b}_0, R, \gamma \rangle$, where $S, A, Z, T, O, \boldsymbol{b}_0$ are defined as in the IO-HMM, $R$ is the reward function so that $R(s,a)$ denotes the immediate reward of taking action $a$ in state $s$, and $\gamma \in [0,1)$ is the discount factor. The goal of an agent is to maximize the expected discounted total reward $\mathbb{E}[\sum_{t=0}^{\infty} \gamma^t R(s_t,a_t)]$ by choosing a policy.

Since the true state is hidden, a policy of agent action must be defined over past actions and observations. If a POMDP is specified, we can use a belief as a sufficient statistic of past actions and observations, where $\pi(\boldsymbol{b},\boldsymbol{a}) = P(a|\boldsymbol{b})$ is a probability of taking action $a$ at belief $\boldsymbol{b}$. A policy $\pi$ induces a value function $V_\pi(\boldsymbol{b})$ that represents the expected discounted total reward of executing policy $\pi$ starting from $\boldsymbol{b}$. It is known (Smallwood & Sondik, 1973) that $V_{\pi^*}$, the value function associated with the optimal greedy policy $\pi^*$, can be approximated with an arbitrary accuracy by a convex, piecewise-linear function

$$Q(\boldsymbol{b},a) = \max_{\boldsymbol{\alpha}\in\Gamma(a)} (\boldsymbol{\alpha}\cdot\boldsymbol{b}) \quad V(\boldsymbol{b}) = \max_a Q(\boldsymbol{b},a) \quad , \quad (2)$$

where $\Gamma(a)$ is a finite set of vectors called $\boldsymbol{\alpha}$-vectors associated to action $a$, and $\boldsymbol{\alpha}\cdot\boldsymbol{b}$ is the inner product of a $\boldsymbol{\alpha}$-vector and vector $\boldsymbol{b}$. We consider soft-max policy for a given set of $\boldsymbol{\alpha}$-vectors:

$$\tilde{\pi}(\boldsymbol{b},a) = \frac{\exp(\beta Q(\boldsymbol{b},a))}{\sum_{a'} \exp(\beta Q(\boldsymbol{b},a'))} \quad (3)$$

where $\beta$ is the inverse temperature parameter that controls the orderedness of the policy. We denote the soft-max policy from the optimal action-value function $Q^*$ as soft-max optimal policy $\tilde{\pi}^*$.

In general, computing an approximately optimal solution within a given error bound $\epsilon$ is NP-hard. However, it is known that given a set of balls of radius $\delta \le O(\epsilon)$ over beliefs that cover an optimal reachable space, an approximated solution can be computed in a polynomial time (Hsu et al., 2008). SARSOP (Kurniawati et al., 2008) is one of approximated POMDP solvers that implements elaborated point selection and a pruning algorithm.

In this paper, we consider situations in which some part of the environment model is uncertain. We introduce a K-element parameter vector $\boldsymbol{\theta}$ with its prior distribution $p(\boldsymbol{\theta})$, and consider a POMDP $\mathcal{P}_{\boldsymbol{\theta}} = \langle S, A, Z, T_{\boldsymbol{\theta}}, O_{\boldsymbol{\theta}}, \boldsymbol{b}_{0,\boldsymbol{\theta}}, R_{\boldsymbol{\theta}}, \gamma\rangle$, where $T$, $O$, $\boldsymbol{b}_0$ and $R$ are determined according to the given parameter $\boldsymbol{\theta}$. An $L$-length sequence $D = (a_1 z_1 \cdots a_L z_L)$ of demonstration by an expert is given, assuming that the expert knows $\boldsymbol{\theta}_{\text{true}}$, the true parameter of the environment, and is following a soft-max optimal policy $\tilde{\pi}^*_{\boldsymbol{\theta}_{\text{true}}}$ under POMDP $\mathcal{P}_{\boldsymbol{\theta}_{\text{true}}}$ with inverse temperature $\beta$.[2] What we want is to calculate $p(\boldsymbol{\theta}|D)$, the posterior distribution of the parameter, and to find an optimal policy over the posterior.

[2]For notational simplicity we assume $\beta$ is known and fixed. However it is easy to apply our methods to cases with unknown $\beta$, because it is equivalent to a fixed $\beta$ with an unknown scaling parameter of the reward function.

## 3. Inferring Posterior

Bayes' theorem gives posterior distribution $p(\boldsymbol{\theta}|D)$ of parameter $\boldsymbol{\theta}$ given demonstration $D = (a_1 z_1 \cdots a_L z_L)$:

$$p(\boldsymbol{\theta}|D) \propto p(D|\boldsymbol{\theta})p(\boldsymbol{\theta}) \quad . \quad (4)$$

Likelihood $p(D|\boldsymbol{\theta})$ of the demonstration is the result of marginalizing expert's policy $\pi$:

$$p(D|\boldsymbol{\theta}) = \int p(D|\boldsymbol{\theta},\pi)p(\pi|\boldsymbol{\theta})d\pi \quad . \quad (5)$$

Note that, from our assumption, the expert's policy $\pi_{\boldsymbol{\theta}}$ is equal to the soft-max optimal policy $\tilde{\pi}^*_{\boldsymbol{\theta}}$ for the POMDP $\mathcal{P}_{\boldsymbol{\theta}}$ with parameter $\boldsymbol{\theta}$, thus $p(\pi = \tilde{\pi}^*_{\boldsymbol{\theta}}|\boldsymbol{\theta}) = 1$.

We can further refactor the likelihood $p(D|\boldsymbol{\theta},\pi_{\boldsymbol{\theta}})$ as follows:

$$\begin{aligned} p(D|\boldsymbol{\theta},\pi) &= p(a_1|\boldsymbol{\theta},\pi)p(z_1|\boldsymbol{\theta},a_1)p(a_2|\boldsymbol{\theta},\pi,a_1 z_1)\cdots \quad (6)\\ &= p(a_1\cdots a_L|\boldsymbol{\theta},\pi,z_1\cdots z_{L-1}) \\ &\quad \cdot p(z_1\cdots z_L|\boldsymbol{\theta},a_1\cdots a_L) \ . \quad (7)\end{aligned}$$

The first factor of Eq. 7 corresponds to the likelihood that the expert performs action $a_i$ given the policy $\pi_{\boldsymbol{\theta}}$:

$$p(a_1\cdots a_L|\pi,z_1\cdots z_{L-1}) = \prod_{i=1}^{L} \pi(\boldsymbol{b}_{i,\boldsymbol{\theta},D},a_i) \quad , \quad (8)$$

where $\boldsymbol{b}_{i,\boldsymbol{\theta},D}$ is the belief at time step $i$ in a POMDP $\mathcal{P}_{\boldsymbol{\theta}}$ with history $D$, calculated by applying Eq. 1 repeatedly to $\boldsymbol{b}_{0,\boldsymbol{\theta}}$.

On the other hand, the second factor corresponds to the likelihood that the environment responds with observation $z_i$ to the performed actions:

$$p(z_1\cdots z_L|\boldsymbol{\theta},a_1\cdots a_L) = \prod_{i=1}^{L}\sum_{s\in S} b_{\boldsymbol{\theta},D,i-1}(s)\sum_{s'\in S} T(s,a_i,s')O(a_i,s',z_i) \ . \quad (9)$$

To our knowledge, previous studies that use the benefit of the first factor in the inference of parameter $\boldsymbol{\theta}$ only consider the change of the reward function. In cases where only the rewards are uncertain, the inference is relatively easy since the value function $V_\pi$ for a given policy is given as a linear function of the reward values (Ramachandran & Amir, 2007). However, if we consider cases where transition and observation probabilities are uncertain, the inference is complex because of the nonlinear dependence between the parameters and the value function.

### 3.1. Maximum A Posteriori Inference

Maximum a posteriori (MAP) inference is to find $\boldsymbol{\theta}$ that maximizes the posterior (Eq. 4). Unfortunately, it is not easy to use sophisticated optimization techniques using gradients because changes in beliefs complicates obtaining gradients for either factor in Eq. 7. This is the major difference from the setting of inverse reinforcement learning, in which we can evaluate the gradient of expert action likelihood, and the observation likelihood is constant given $D$.

We take a straightforward approach to optimization by using the COBYLA algorithm (Powell, 1998), which does not require gradients. For each candidate parameter value $\boldsymbol{\theta}$, we call a POMDP solver for POMDP $\mathcal{P}_{\boldsymbol{\theta}}$ to obtain the optimal action-value function $Q^*$, which gives the soft-max optimal policy $\tilde{\pi}^*_{\boldsymbol{\theta}}$ for the POMDP that is used for evaluating expert action likelihood (Eq. 8). As for the observation likelihood (Eq. 9), we apply the standard forward algorithm for IO-HMM to POMDP $\mathcal{P}_{\boldsymbol{\theta}}$ and sequence $D$.

This algorithm has no guarantee to find the MAP parameter because it is based on a local search. However, in practice it seems to find a reasonably good solution, and calculation is quick compared to the sampling approach which we will describe next.

### 3.2. Inference by Sampling

We also employ a Markov chain Monte Carlo (MCMC) sampling approach (Gilks et al., 1996) to infer the posterior distribution of $\boldsymbol{\theta}$. Unlike MAP inference, The approximation calculated by MCMC can be arbitrarily accurate with a sufficient computational time.

The traditional way of sampling parameters for IO-HMM is to use the Markov chain Monte Carlo approach; that is by alternately sampling the hidden state sequence $\boldsymbol{s}$ given parameter $\boldsymbol{\theta}$, and $\boldsymbol{\theta}$ given $\boldsymbol{s}$. By using a conjugate prior for the parameters, we can easily sample $\boldsymbol{\theta}$ from the posterior given $\boldsymbol{s}$.

To make the sample distribution follow the expert action likelihood, we introduce the Metropolis algorithm (Metropolis et al., 1953), which accepts the proposed sample $\boldsymbol{\theta}'$ with the probability $\min(1, p'/p)$, where $p'$ and $p$ are the expert action likelihood for the proposed sample and for the previous sample, respectively.

**Algorithm 1** MCMC sampler for posterior $p(\boldsymbol{\theta}|D)$

**Require:** $D$: demonstration
1: sample $\boldsymbol{\theta}$ from the prior
2: $p :=$ infinitesimal positive value
3: **loop**
4: sample $\boldsymbol{s} = (s_1 \cdots s_L)$ from $p(\boldsymbol{s}|D, \boldsymbol{\theta})$
5: **for** $k := 1$ **to** $K$, in random order **do**
6: $\boldsymbol{\theta}' := \boldsymbol{\theta}$
7: replace $\theta'_k$ by a sampled value from IO-HMM posterior given $\boldsymbol{s}, D$
8: call POMDP solver to find $\tilde{\pi}^*_{\boldsymbol{\theta}'}$
9: $p' := p(a_1 \cdots a_L | \tilde{\pi}^*_{\boldsymbol{\theta}'}, z_1 \cdots z_L)$ {Eq. 8}
10: **if** with probability $\min(1, p'/p)$ **then**
11: $p := p'$, $\boldsymbol{\theta} := \boldsymbol{\theta}'$ {accept the sample}
12: **end if**
13: **end for**
14: **end loop**

Algorithm 1 shows the sampler for the posterior of model parameters. The algorithm is similar to the sampling algorithm for model parameters from IO-HMM posterior, which deals with only the likelihood of an environmental response (the second term in Eq. 7). The difference lies in lines 8–12, that performs Metropolis algorithm for the expert action likelihood (the first term in Eq. 7). We run the algorithm until specified number $M$ of samples are collected, excluding burn-ins and interval samples.

## 4. Planning with a Sampled Posterior

Our goal is to achieve model-parameter apprenticeship learning; that is, to make an optimal policy for the learned posterior of POMDP model parameters. In this section, we describe how to produce an optimal policy based on the sampled results. Note that, in case of a MAP estimate, we can obtain a policy by applying a solver to POMDP $\mathcal{P}_{\hat{\boldsymbol{\theta}}}$ with estimated parameter $\hat{\boldsymbol{\theta}}$.

Existing planning methods for POMDP with Bayesian uncertainty (Ross et al., 2008) are not applicable, because they require that the uncertainty be represented in conjugate priors, which cannot represent the posterior distribution of parameters after observing demonstration. Instead, we employed a method to develop a POMDP policy based on the sampled parameters. The idea is to extend the hidden state of POMDP with a variable $m$, which is an index of the sampled parameters $\boldsymbol{\theta}_m$ $(m = 1, \ldots, M)$. At the beginning $m$ is uniformly distributed, and never changes. This extended POMDP can be solved by a standard POMDP solver. We expect that the belief over sample index $m$ converges to the index of the most likely parame-

*Table 1.* Distribution of the estimated posterior parameters.

| | | Error of prior mean | IO-HMM Sampler | IO-HMM EM | Proposed Sampler | Proposed MAP |
|---|---|---|---|---|---|---|
| $p_i$: prob. of tiger position | mean error | -0.100 | -0.039 | -0.050 | -0.009 | -0.007 |
| | RMSE | | 0.143 | 0.091 | 0.057 | 0.059 |
| | s.d. samples | | 0.110 | | 0.059 | |
| $p_l$: prob. of hear left when the tiger is left | mean error | -0.183 | -0.151 | -0.066 | -0.034 | -0.034 |
| | RMSE | | 0.189 | 0.145 | 0.047 | 0.048 |
| | s.d. samples | | 0.080 | | 0.040 | |
| $p_r$: prob. of hear right when the tiger is right | mean error | -0.183 | -0.193 | -0.104 | -0.042 | -0.043 |
| | RMSE | | 0.241 | 0.170 | 0.057 | 0.061 |
| | s.d. samples | | 0.090 | | 0.046 | |
| $r_t$: reward of seeing the tiger | mean error | 50.000 | — | — | 13.548 | 17.514 |
| | RMSE | | — | — | 19.356 | 22.336 |
| | s.d. samples | | — | | 21.812 | |

RMSE: Root mean squared error of the estimate values.
s.d. samples: Average standard deviation of sampled values.

ter while the agent interacts with the environment. In case the target POMDP is episodic, we want to retain belief over $m$ beyond episodes, so we convert the target POMDP into non-episodic POMDPs before extension.

Formally, given $M$ sampled parameters $\boldsymbol{\theta}_1, \ldots, \boldsymbol{\theta}_M$ for the target POMDP, we create an extended POMDP $\tilde{\mathcal{P}} = \langle \tilde{S}, A, Z, \tilde{T}, \tilde{O}, \tilde{\boldsymbol{b}}_0, \tilde{R}, \gamma \rangle$, where

$$\tilde{S} = S \times \{1, \ldots, M\} \qquad \tilde{O}([s,m], a, z) = O_{\boldsymbol{\theta}_m}(s, a, z)$$

$$\tilde{b}_0([s,m]) = b_{0,\boldsymbol{\theta}_m}(s)/M \qquad \tilde{R}([s,m], a) = R_{\boldsymbol{\theta}_m}(s, a)$$

$$\tilde{T}([s,m], a, [s',m']) = \begin{cases} T_{\boldsymbol{\theta}_m}(s, a, s') & m = m' \\ 0 & m \neq m' \end{cases}.$$

Note that the optimal policy of the extended POMDP becomes a good policy in the target POMDP only if the samples represent the target well. If we need a agent that learns by exploring the uncertainty in the target POMDP, we will need scheduled resampling as has been done in fully observable environments by the BOSS algorithm (Asmuth et al., 2009). In this paper we chose not to resample, because our purpose is to evaluate the posterior distribution $p(\boldsymbol{\theta}|D)$ obtained from demonstration.

## 5. Experiments

To evaluate the proposed model-parameter apprenticeship learning algorithms, we performed experiments with two tasks: one is a simple environment based on the well-known Tiger problem (Kaelbling et al., 1998), and the other is a task designed for a dialog system. In the following experiments, we used APPL Toolkit which implements the SARSOP algorithm (Kurniawati et al., 2008) as a POMDP solver. We used COBYLA implementation in the NLopt library (Johnson, 2008).

### 5.1. Bayesian Tiger Problem

We introduced four unknown parameters to the Tiger problem, whose prior is represented as $p_i \sim \mathrm{Beta}(3,3)$, $p_l, p_r \sim \mathrm{Beta}(5,3)$, $r_t \sim \mathcal{N}(-50, 50^2)$ as follows. An agent is standing in front of two doors. A tiger is hidden behind the left door with probability $p_i$ and behind the right door with probability $1 - p_i$. The agent can open one of the doors, and obtain reward $r_t$ if the agent sees the tiger and reward 10 otherwise. Alternatively, the agent can choose to listen with reward $-1$: if the tiger is behind the left door, the agent hears the tiger from the left with probability $p_l$ and from the right with probability $1 - p_l$; if the tiger is behind the right door, the agent hears it from the right with probability $p_r$ and from the left with probability $1 - p_r$.

We set the true environment as $p_i = 0.6$, $p_l = p_r = 0.85$ and $r_t = -100$, and we used $\gamma = 0.9$. We generated 100 demonstrations by the experts with soft-max policy $\beta = 0.3$, each consisting of 100 steps of actions and observations (which contained 22 episodes on average). For each demonstration, we applied one of the learning algorithms to the demonstration to estimate the posterior. From the estimated posterior, an optimal greedy policy was derived, and tested by simulating 100,000 steps in the true environment, and the average reward was measured. For sampling algorithms, 1,000 MCMC steps were performed including 100-step burn-in, and parameters were sampled for every 10 steps (total $M = 90$ samples) to generate a greedy policy.

Table 1 shows the distribution of the estimated parameters. Both of the proposed methods produce more accurate estimates of parameters compared to the meth-ods based on IO-HMM[3]. We can see that the proposed methods provide better RMSE than do the IO-HMM methods. The estimates from IO-HMM methods are closer to the prior mean, suggesting that the provided demonstration is too short to obtain an accurate estimate. On the other hand, the estimates from the proposed methods are closer to the true value, which indicates that the proposed methods provide a better estimate using the same length of demonstration. We can also see that the proposed sampler produces a narrower posterior distribution (i.e., smaller standard deviation of the samples) than that of the IO-HMM sampler.

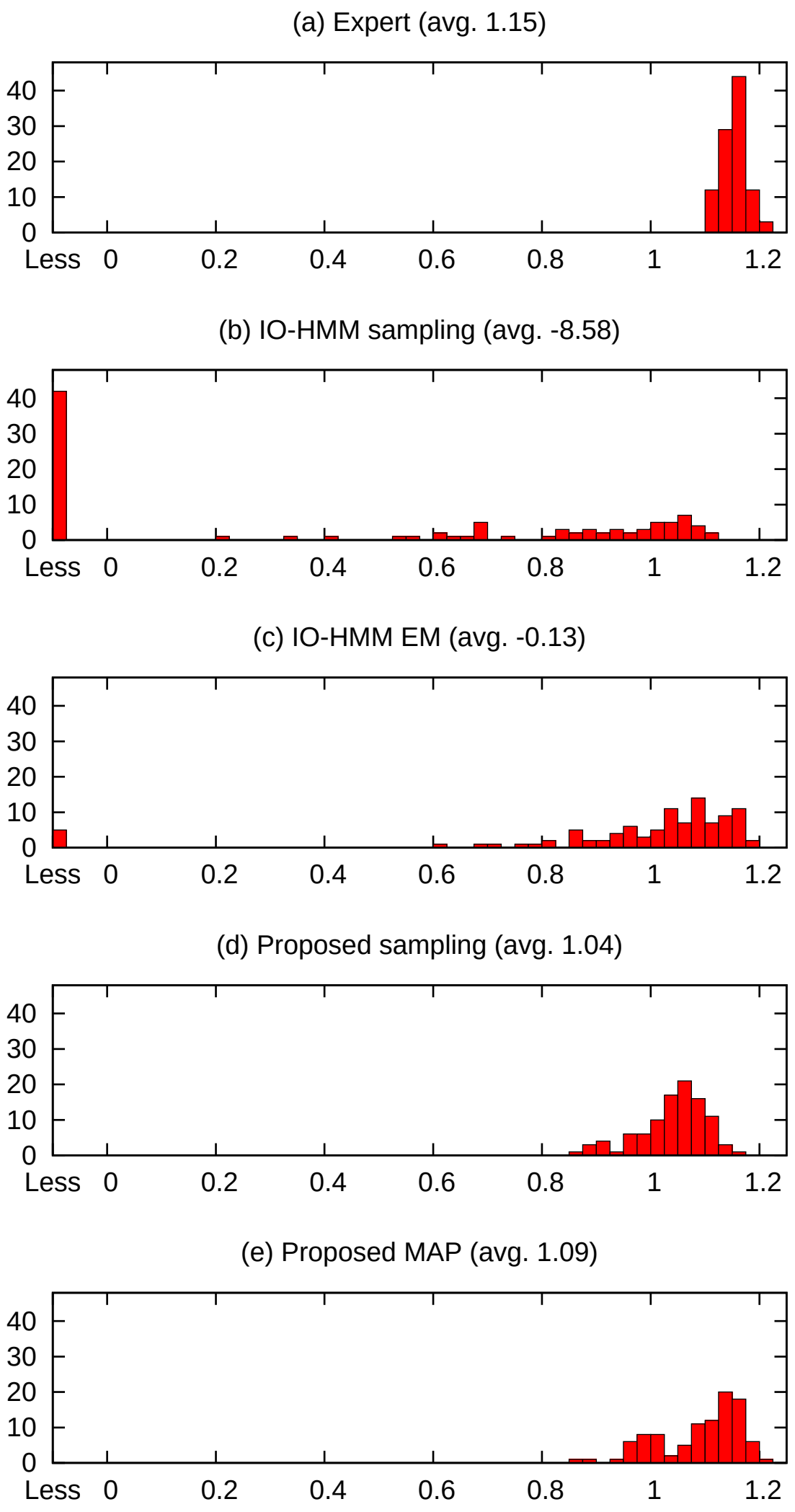

*Figure 2.* Histogram of average reward obtained by simulating the learned greedy policy. The leftmost bar ("Less") is the sum of counts that falls off the range of these plots.

As shown in Figure 2, having an accurate estimate leads to better results in simulation by the learned policy. The results of the policies based on estimated posterior with our methods are not much worse than those of the expert policy who knows the true parameter values. On the other hand, policies based on IO-HMM estimation occasionally result in very bad policies, as shown in "Less" average rewards in the figure. Considering that the demonstration is short and noisy, these results indicate that the model-parameter apprenticeship learning methods prevent agents from critical failures in learning to follow the demonstrated task.

### 5.2. Dialog System

To show the effectiveness of our methods in a more realistic scenario, we developed a new task of dialog management for a ticket-vending system. A user asks the agent for a ticket with a certain origin and destination via an unreliable voice recognition interface; the task of the agent is to repeat the order correctly, and issue the ticket. We expect that the expert demonstration is useful to determine parameters, which represents user's preferred ticket routes and way of talking.

The task consists of 13 observations from voice recognition, including three place names and SIL (silence). The agent can choose from 11 actions, consisting of uttering one of 9 words, waiting for next word from the user, or issuing a ticket. The dialog is managed by a 32-state POMDP (Fig. 3) for each of $3 \times 2 = 6$ ticket routes, resulting in the total of 192 hidden states.

The POMDP is parameterized with a 15-dimensional vector $\boldsymbol{\theta}$; 4 parameters are assigned to route preferences (initial state distribution), 9 to ways of talking (transition probabilities), and 2 to voice recognition errors (observation probabilities). The agents are required to estimate the parameters from a 300-step demonstration generated by an expert. In the experiments we didn't use samplers since they require too much computational resources.

We generated 12 demonstrations by the experts (the result of solving the true POMDP model) with softmax policy $\beta = 0.4$, each consisting of 300 steps of actions and observations (which contained 27 episodes on average). Using the learned parameters, we applied SARSOP POMDP solver to obtain a greedy policy, and measured average reward by testing the policy on the original environment. Since calculating the exact solution of the POMDP is too expensive, we set the timeout of 40 CPU seconds for each parameter can-didate during MAP search, and 600 CPU seconds for calculating the expert policy and the final policy based on the estimated parameters;

[3]Note that we compare only state transition parameters because the rewards cannot be estimated by IO-HMM methods.

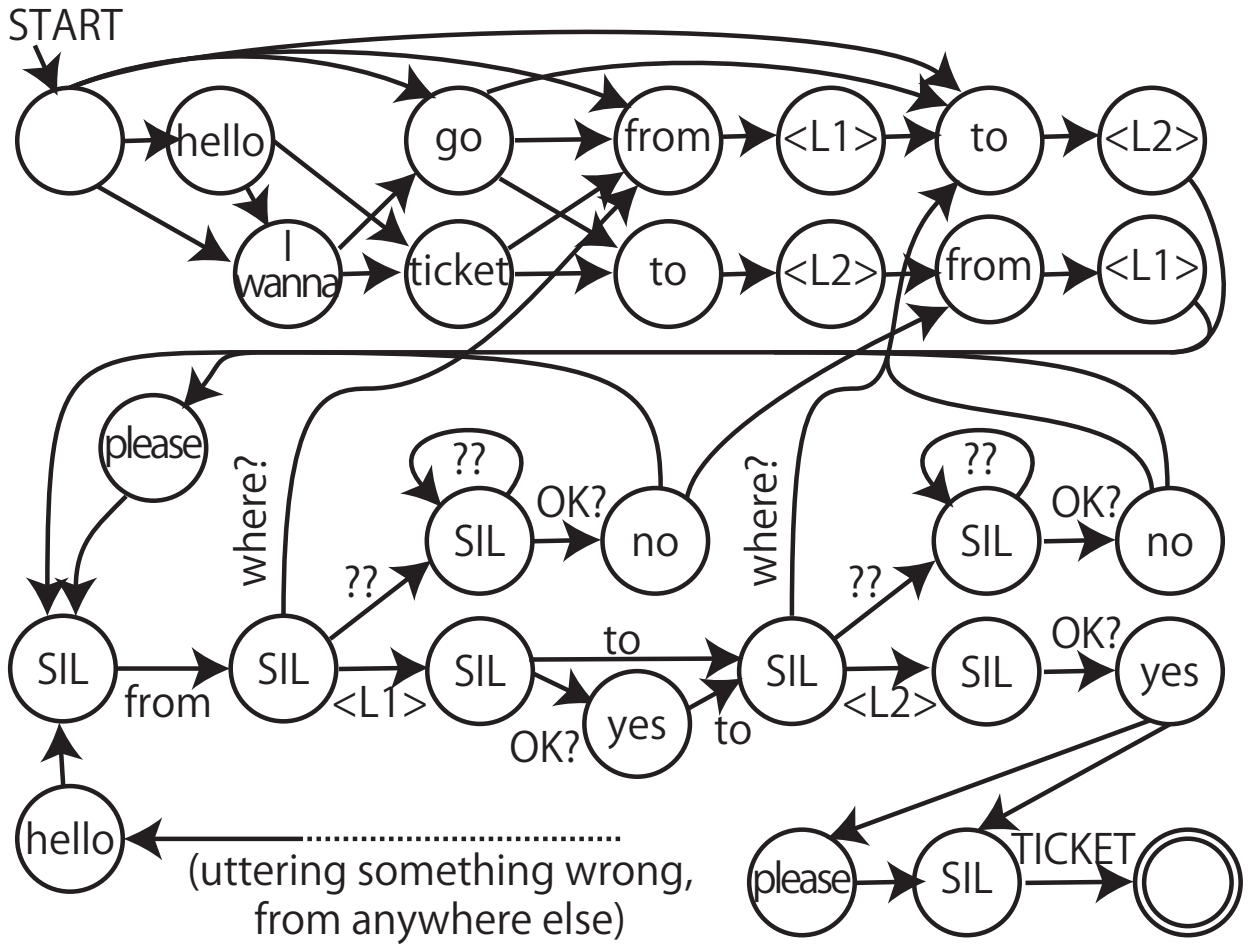

*Figure 3.* Dialog System task. A node label denotes a user voice (observation), and an edge label denotes a system utterance (action). <L1> and <L2> denote origin and destination places, respectively (depending on hidden user state). For brevity, rewards, probabilities and some transitions/states are omitted.

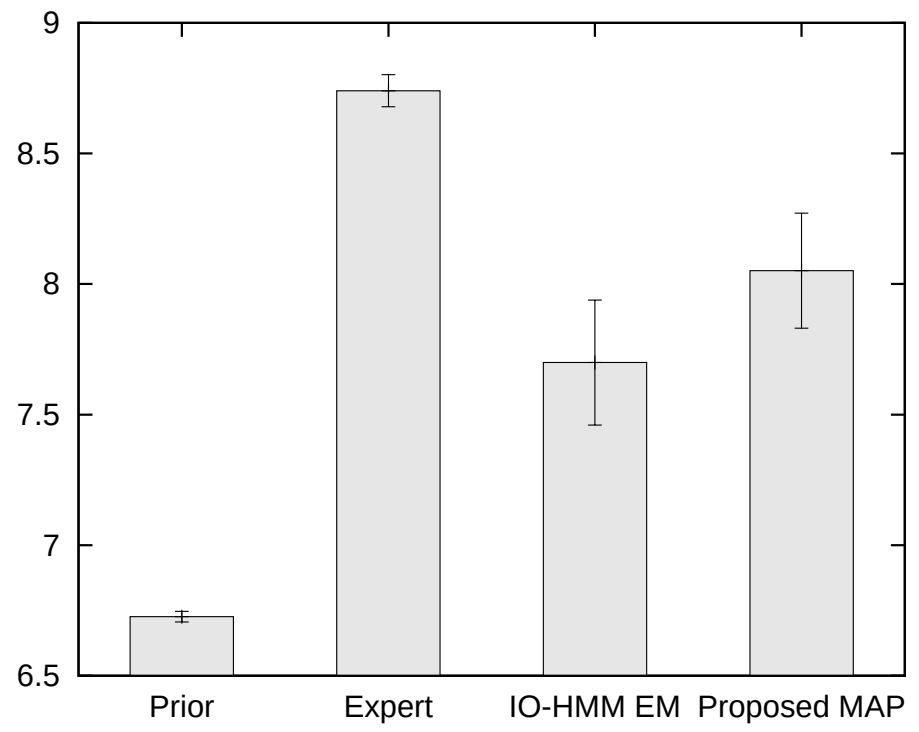

*Figure 4.* The average rewards obtained by the learned greedy policy of the dialog task. The error bars show $p < 0.05$ confidence interval.

Figure 4 shows the results. We found that the agents based on the parameters estimated by the proposed MAP algorithm perform significantly better than the agents based on the parameters estimated by IO-HMM ($P < .05$). However, in this setting, we couldn't obtain the expert-level performance by the apprenticeship learning. One possible reason is that the optimization algorithm is disturbed by the approximation error of expert action likelihood, which is caused by the short timeout of the POMDP solver and random searching strategy of SARSOP. We believe that the result can be improved if we use more computational resources; or, if we use a POMDP solver that can be started from the result of a similar POMDP, we may be able to improve the optimization process.

## 6. Conclusion

We have shown that the apprenticeship learning approach can be used to estimate parameters of an unknown POMDP environment. Assuming that an expert knowing the perfect POMDP model of the target environment will try to maximize the reward, we can extract the expert's knowledge about the environment from his demonstration in terms of the posterior distribution of unknown parameters. Our proposed algorithms are simple but are capable of estimating POMDP parameters accurately even if the demonstration is short. We also showed that the extracted knowledge can be used to develop a policy that can act reasonably well in the target environment.

Our approach is a generalization of inverse reinforcement learning, in a sense that the unknown parameters are not limited to those for the reward function but can also be for transition and observation functions. This approach can be particularly useful in the domain of applications that interact with human beings, whose model is unknown but demonstration by experts is available. One direct extension of the approach is to estimate other parameters, such as the discount factor of the expert, from the demonstration. Future work should also include the development of more efficient algorithms as has been done in the context of inverse reinforcement learning.

## Acknowledgment

This research is supported by the Aihara Innovative Mathematical Modelling Project, the Japan Society for the Promotion of Science (JSPS) through the "Funding Program for World-Leading Innovative R&D on Science and Technology (FIRST Program)," initiated by the Council for Science and Technology Policy (CSTP), and by JSPS Grant-in-Aid for Young Scientists (B) (20700126).